\documentclass{article}

\usepackage[dblblindworkshop,final]{neurips_2026}
\usepackage[utf8]{inputenc}
\usepackage[T1]{fontenc}
\usepackage{hyperref}
\usepackage{url}
\usepackage{booktabs}
\usepackage{array}
\usepackage{amsfonts}
\usepackage{amsmath}
\usepackage{graphicx}
\usepackage{placeins}
\usepackage{microtype}
\usepackage{xcolor}
\usepackage{orcidlink}
\hypersetup{
  pdftitle={QuantCortex: Auditing Financial ML Backtests with Evaluation Contracts and Executable Checks},
  pdfauthor={Jimmy Chan and Kevin Lee},
  colorlinks=false,
  pdfborder={0 0 1},
  linkbordercolor={1 0 0},
  citebordercolor={0 1 0},
  urlbordercolor={0 1 1}
}

\graphicspath{{figures/}}
\newcommand{\system}{QuantCortex}
\newcommand{\PaperNetCAGR}{2.05\%}
\newcommand{\PaperGrossCAGR}{3.86\%}
\newcommand{\PaperCashCAGR}{2.50\%}
\newcommand{\PaperMatchedCAGR}{7.11\%}
\newcommand{\PaperCostedComparatorCAGR}{6.10\%}
\newcommand{\PaperZeroCashCAGR}{0.47\%}
\newcommand{\PaperNetSharpe}{\mbox{-0.04}}
\newcommand{\PaperGrossSharpe}{0.25}
\newcommand{\PaperMatchedSharpe}{0.85}
\newcommand{\PaperCostedComparatorSharpe}{0.67}
\newcommand{\PaperZeroCashSharpe}{0.11}
\newcommand{\PaperNetMaxDrawdown}{\mbox{-9.25\%}}
\newcommand{\PaperGrossMaxDrawdown}{\mbox{-8.51\%}}
\newcommand{\PaperCashMaxDrawdown}{\mbox{-0.45\%}}
\newcommand{\PaperMatchedMaxDrawdown}{\mbox{-6.21\%}}
\newcommand{\PaperCostedComparatorMaxDrawdown}{\mbox{-6.22\%}}

\newcommand{\PaperZeroCashMaxDrawdown}{\mbox{-10.59\%}}

\newcommand{\PaperFullyCash}{46.89\%}

\newcommand{\PaperCostedActiveMean}{\mbox{-3.86\%}}
\newcommand{\PaperCostedActiveLower}{\mbox{-6.40\%}}
\newcommand{\PaperCostedActiveUpper}{\mbox{-1.35\%}}

\newcommand{\PaperZeroCashSHVSharpe}{\mbox{-0.31}}

\newcommand{\PaperDecompositionRows}{%
  Active risky allocation & \mbox{-3.05\%} & [\mbox{-5.52\%}, \mbox{-0.57\%}] & 35/5000 \\
  Dynamic exposure timing & 0.54\% & [\mbox{-2.35\%}, 3.45\%] & 3182/5000 \\
  Passive risky exposure & 4.00\% & [0.85\%, 7.04\%] & 4963/5000 \\
  Modeled implementation cost & \mbox{-1.76\%} & [\mbox{-2.12\%}, \mbox{-1.41\%}] & 0/5000 \\
  Net excess return over cash & \mbox{-0.26\%} & [\mbox{-4.29\%}, 3.82\%] & 2340/5000 \\
}
\newcommand{\PaperScoreCorrectionRows}{%
  Original score & 2.05\% & \mbox{-0.044} & 30.97\% \\
  Corrected score & 2.53\% & 0.032 & 27.91\% \\
}

\newcommand{\ExpansionEvaluationSessions}{2,011}
\newcommand{\ExpansionBootstrapReplications}{5,000}

\newcommand{\ExpansionBaselineRows}{\shortstack[l]{United States\\sectors} & Time series momentum & 11.44\% & 0.49 & 10.29\% & -34.16\% \\
\shortstack[l]{United States\\sectors} & Cross-sectional momentum & \textbf{14.19\%} & \textbf{0.60} & \textbf{13.04\%} & \textbf{-32.06\%} \\
\shortstack[l]{United States\\sectors} & Short-term reversal & 11.02\% & 0.44 & 9.54\% & -45.08\% \\
\shortstack[l]{United States\\sectors} & Gradient boosting & 13.66\% & 0.53 & 12.10\% & -39.96\% \\
Country equities & Time series momentum & 2.18\% & -0.02 & 0.88\% & -40.45\% \\
Country equities & Cross-sectional momentum & \textbf{8.84\%} & \textbf{0.33} & \textbf{7.21\%} & -40.73\% \\
Country equities & Short-term reversal & 6.03\% & 0.20 & 4.48\% & -43.91\% \\
Country equities & Gradient boosting & 7.87\% & 0.31 & 6.60\% & \textbf{-33.56\%} \\}
\newcommand{\ExpansionTradingRows}{\shortstack[l]{United States\\sectors} & Time series momentum & 99.90\% & 4.39 & 0.57\% \\
\shortstack[l]{United States\\sectors} & Cross-sectional momentum & 99.90\% & 5.26 & 0.68\% \\
\shortstack[l]{United States\\sectors} & Short-term reversal & 77.73\% & 13.79 & 1.80\% \\
\shortstack[l]{United States\\sectors} & Gradient boosting & 98.74\% & 12.64 & 1.65\% \\
Country equities & Time series momentum & 88.61\% & 4.89 & 0.64\% \\
Country equities & Cross-sectional momentum & 99.90\% & 5.20 & 0.68\% \\
Country equities & Short-term reversal & 70.88\% & 13.91 & 1.81\% \\
Country equities & Gradient boosting & 89.26\% & 13.13 & 1.71\% \\}
\newcommand{\ExpansionRankRows}{\multicolumn{7}{@{}l@{}}{\textit{United States sectors}} \\
Time series momentum & $0.494$ & 3 & $0.526$ & 2 & $0.525$ & 4 \\
Cross-sectional momentum & $\mathbf{0.596}$ & 1 & $\mathbf{0.572}$ & 1 & $\mathbf{0.630}$ & 1 \\
Short-term reversal & $0.439$ & 4 & $0.055$ & 4 & $0.531$ & 3 \\
Gradient boosting & $0.529$ & 2 & $0.409$ & 3 & $0.606$ & 2 \\
\addlinespace
\multicolumn{7}{@{}l@{}}{\textit{Country equities}} \\
Time series momentum & $-0.018$ & 4 & $0.044$ & 3 & $0.022$ & 4 \\
Cross-sectional momentum & $\mathbf{0.329}$ & 1 & $\mathbf{0.313}$ & 1 & $0.364$ & 2 \\
Short-term reversal & $0.197$ & 3 & $-0.186$ & 4 & $0.297$ & 3 \\
Gradient boosting & $0.314$ & 2 & $0.244$ & 2 & $\mathbf{0.413}$ & 1 \\}
\newcommand{\ExpansionEffectRows}{\multicolumn{4}{@{}l@{}}{Retroactive assignment minus execution at the first later close} \\
Mean return difference & 3/5/0 & 3/5/0 & 3/5/0 \\
Sharpe difference & 3/5/0 & 3/5/0 & 3/5/0 \\
\addlinespace
\multicolumn{4}{@{}l@{}}{Zero cash return minus SHV cash return} \\
Mean return difference & 5/3/0 & 5/3/0 & 4/4/0 \\
Sharpe difference & 5/3/0 & 5/3/0 & 4/4/0 \\
\addlinespace
\multicolumn{4}{@{}l@{}}{Zero trading cost minus a charge of 13 basis points} \\
Mean return difference & 0/0/8 & 0/0/8 & 0/0/8 \\
Sharpe difference & 0/0/8 & 0/0/8 & 0/0/8 \\
\addlinespace
\multicolumn{4}{@{}l@{}}{Net strategy minus comparison portfolio} \\
Mean return difference & 1/7/0 & 1/7/0 & 1/7/0 \\
Sharpe difference & 1/7/0 & 1/7/0 & 1/7/0 \\
\addlinespace
\multicolumn{4}{@{}l@{}}{Vectorized engine minus event-driven engine} \\
Mean return difference & 0/6/2 & 0/4/4 & 0/4/4 \\
Sharpe difference & 0/7/1 & 0/4/4 & 0/4/4 \\}
\newcommand{\ExpansionCostEffectRange}{0.57 to 1.81 percentage points}

\newcommand{\ExpansionEngineMaxWealthGap}{3.89\%}

\title{QuantCortex: Auditing Financial ML Backtests with Evaluation Contracts and Executable Checks}

\workshoptitle{TAE (Trust-AI-Eval): Can We Trust AI Evaluation?}

\author{%
  Jimmy Chan\,{\hypersetup{pdfborder={0 0 0}}\orcidlink{0009-0002-4670-2828}}\thanks{Equal contribution.} \\
  Pepperdine University \\
  Los Angeles, CA, USA \\
  \texttt{jimmybojim.chan@alumnimail.pepperdine.edu} \\
  \And
  Kevin Lee\,{\hypersetup{pdfborder={0 0 0}}\orcidlink{0009-0004-0388-9260}}$^{*}$ \\
  University of California, Los Angeles \\
  Los Angeles, CA, USA \\
  \texttt{kevinlee69720@g.ucla.edu}
}

\begin{document}

\maketitle
  \begingroup
    \renewcommand{\thefootnote}{\fnsymbol{footnote}}%
    \footnotetext[2]{Available at: \url{https://github.com/magnaprog/quantcortex}.}%
  \endgroup

\begin{abstract}
Portfolio backtests turn model decisions into returns through assumptions about trading, cash income, and costs. We examine how these assumptions shape model comparisons while using the same prices and portfolio decisions in each accounting comparison. \system{}{$^{\dagger}$}{} separates portfolio construction from execution and valuation. Under baseline accounting, new holdings begin earning returns after execution. We study a fund rotation strategy and compare gradient boosting with simple trading rules across sector and country funds. Both studies revisit market histories available during development. Crediting a new allocation with returns from before execution can improve or worsen apparent performance and reverse strategy rankings. Cash income, trading costs, and the model seed also influence the comparisons. The rotation strategy's arithmetic decomposition traces its shortfall in mean return relative to cash to allocation and trading costs that outweigh positive contributions from passive exposure and exposure timing. Resampling consecutive dates together gives pointwise intervals for accounting effects, conditional on each study's models, portfolio decisions, and seeds. Interpreting a model's performance begins with the rules that translate its decisions into holdings and returns.
\end{abstract}

\section{Introduction}

Consider a model that chooses a portfolio after observing today's close. The new holdings can earn returns after execution. Crediting them with today's return assigns the decision an outcome already known when it was made. Cash treatment and trading charges also enter the return calculation. These choices make the translation from decisions to holdings and cash flows part of the model comparison.

We study allocations to exchange-traded funds, or ETFs. A target weight is the fraction of capital a strategy intends to invest in an asset. The backtest translates those intentions into holdings, values the holdings as prices change, and accounts for cash income and trading costs. These calculations determine the return series on which inference rests. Strategy search adds a separate source of uncertainty: repeatedly trying models on the same market history can favor a model that happened to perform well in that sample
\citep{white2000reality,harvey2016crosssection,bailey2014deflated,
bailey2017pbo,arnott2019protocol}.

Our question is how much a model comparison changes when the portfolio decisions stay the same but their accounting changes. We define an evaluation contract as the timing, cash, cost, and accounting rules under which those decisions earn returns, together with checks on their implementation. A record of decision times, asset symbols, and target weights lets us evaluate the same allocations under alternative conventions. A return decomposition then separates contributions from allocation, exposure timing, passive exposure, and trading costs.

We examine this question in two settings. A weekly ETF rotation strategy that predates the audit provides a retrospective case. A second study compares gradient boosting with time series momentum, cross-sectional momentum, and short-term reversal on two ETF sets under a protocol written before data retrieval. Repeating the regression model with five random seeds shows how its portfolio performance varies within that specification.

The comparisons trace performance from the information available at a decision to execution and the resulting returns. Checks on information availability and portfolio accounting support this interpretation. Separate tests simulate stale state and uncertain submission outcomes to examine how the implementation handles them before an order can be retried.

\section{Related work}

\paragraph{Evaluation validity:}
A benchmark measures a particular task, so its connection to a broader capability is part of the evaluation argument \citep{raji2021benchmark}. Reusing a test set adaptively changes the inferential problem \citep{dwork2015reusable}, while leakage can make information available during evaluation that would be unavailable at prediction time \citep{kapoor2023leakage}. Financial backtests add an accounting step: portfolio decisions become returns through execution, valuation, and cash flows. We examine that step under explicit rules.

\paragraph{Financial backtest validity:}
Sharpe inference depends on assumptions about distributions and serial dependence
\citep{lo2002statistics}. Searching many strategies on the same history can overstate the performance of the one ultimately chosen
\citep{white2000reality,bailey2014deflated,bailey2017pbo,harvey2016crosssection}.
\citet{arnott2019protocol} organize these risks into a research protocol, and
\citet{zhang2026alpha} study contamination at decision time across models and equity datasets. We examine a timing error in which a portfolio decision receives the return from an interval that ends before execution is possible.

\paragraph{Attribution and comparison:}
Performance analysis has long separated decisions into return components
\citep{fama1972components,brinson1985measuring}. Our decomposition accounts for daily return above cash through passive exposure, ex post exposure timing, active allocation, and trading costs. Allocation and exposure are measured relative to a reference basket. The portfolio rules use momentum,
reversal, residual momentum, and gradient boosting
\citep{jegadeesh1993returns,moskowitz2012timeseries,lehmann1990fads,
blitz2011residual,friedman2001greedy}.
We adapt the momentum and reversal rules to ETF portfolios that hold positive positions or cash. Section~\ref{sec:portfolio-study} gives their windows and weights. The retrospective rule scores ETF residuals from a
regression on one factor. Blitz et al. construct stock portfolios from residuals of the Fama and French model with three factors: market, size, and value. Their study motivates the use of residual signals. Our ETF universe, regression on one factor, and allocation rule differ from theirs.

\paragraph{Checks and reproducibility:}
CheckList tests natural language processing behaviors beyond aggregate accuracy \citep{ribeiro2020checklist}; our checks address the financial evaluation rules defined here. Qlib and FinRL provide modular workflows for financial AI \citep{yang2020qlib,liu2020finrl}, while FinRL-X joins data, strategy construction, backtesting, and broker execution through a weight interface \citep{yang2026finrlx}. Reproducing submitted financial analyses from the same raw inputs is difficult \citep{perignon2024computational}. The NeurIPS 2019 reproducibility program combined code, community reproduction, and a checklist \citep{pineau2021reproducibility}.

\citet{yin2026implementation} compare five engines across 15 strategies. We examine what changes when two engines process the same allocation history. Their conventions differ in daily weight resets, drift between trades, trade accounting, and sizing after costs, so the comparison measures their combined effect.

\section{Evaluation contract}

\begin{figure}[t]
  \centering
  \includegraphics[width=\textwidth]{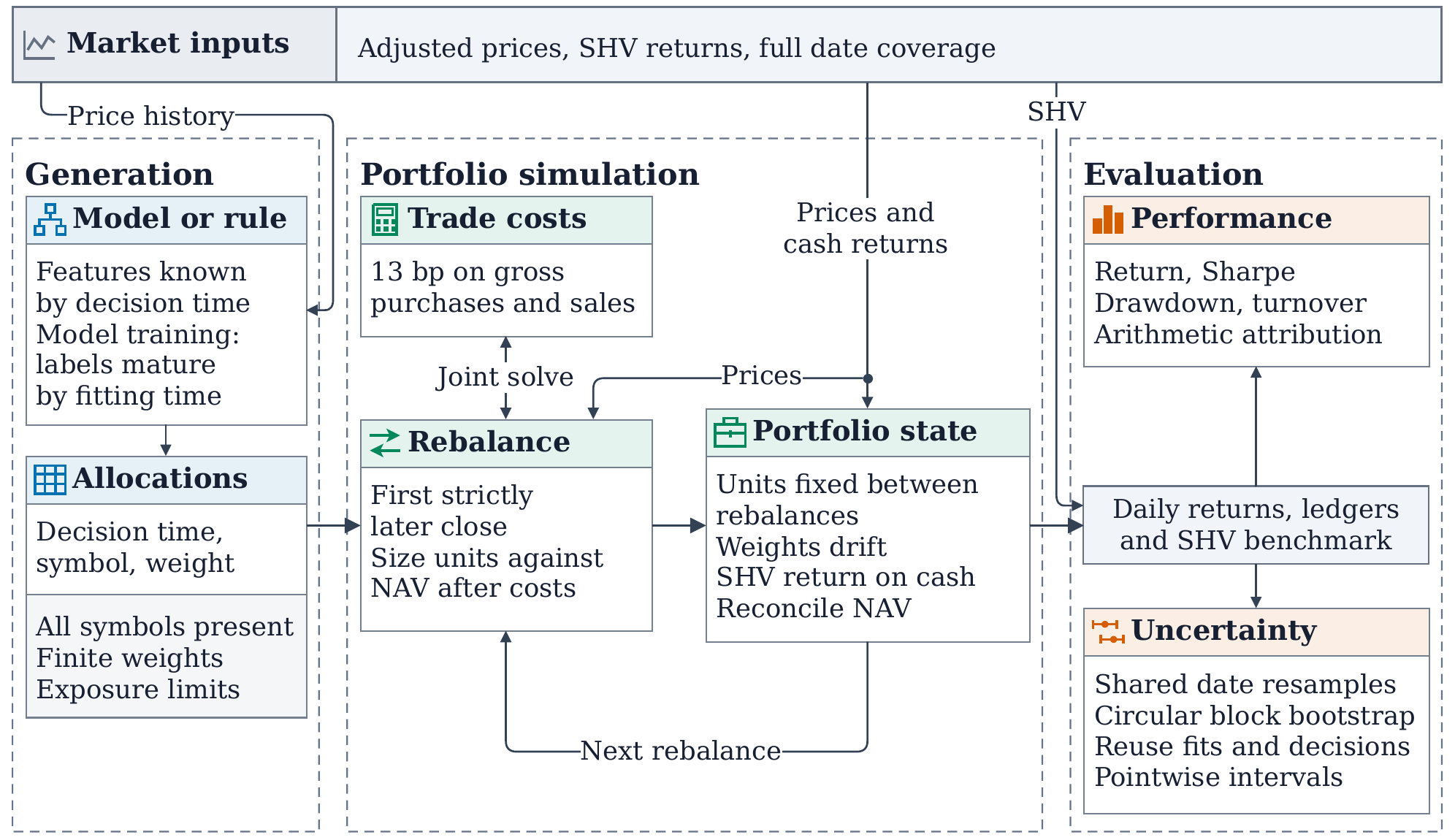}
  \caption{Baseline chronology and accounting. Portfolio generation uses features available by decision time and training outcomes available by fitting time. The event engine solves for trade quantities and costs together, deducts costs from portfolio value at execution, and credits the new holdings with returns over the next interval. Timing, cash, and cost comparisons reuse the same portfolio decisions. Each return pair and SHV use the same dates for metric calculation and circular block resampling. Recomputing metrics across resamples gives pointwise intervals conditional on the original models, portfolio decisions, and seeds.}
  \label{fig:protocol}
\end{figure}

A valid return calculation follows the portfolio's chronology and reconciles its wealth. We apply design by contract to these portfolio invariants \citep{meyer1992contract}: a run stops when an input or accounting quantity violates one of these conditions. Together, the rules determine holdings, cash, costs, and returns. Figure~\ref{fig:protocol} follows the calculation, and Table~\ref{tab:invariants} connects each rule to its check or supporting record.

\begin{table}[t]
  \caption{Evaluation rules and the checks or records used to assess them.}
  \label{tab:invariants}
  \centering
  \begin{tabular}{>{\raggedright\arraybackslash}p{0.16\linewidth}
                  >{\raggedright\arraybackslash}p{0.40\linewidth}
                  >{\raggedright\arraybackslash}p{0.34\linewidth}}
    \toprule
    Area & Rule & Check or evidence \\
    \midrule
    Allocation record & Decision timestamp, symbol, and target weight; each decision gives a weight for every asset, with nonnegative weights within the gross limit. & Schema and runtime checks reject records with invalid fields, incomplete books, and invalid weights. \\
    Information and timing & Features must be available by their decision times, and training labels when fitting begins. Execution uses the first later bar. & The generator checks feature and label availability. The evaluator checks execution alignment. \\
    Cash and cost & Residual cash earns the return calculated from adjusted closing prices of the Treasury ETF SHV; 13 basis points apply to the value of purchases plus sales. & Checks reject missing cash returns and absent cost models; ledgers reconcile. \\
    Comparison & Basket return at actual exposure; a separate portfolio follows target exposure and pays its own costs. & Daily return reconciliation, reproducible comparisons, and cost records. \\
    Engines & The event engine changes risky asset quantities only when it trades; the vectorized engine resets weights daily. & Common target weights reveal differences between the accounting conventions. \\
    Order state & Record intent before submission. After an uncertain outcome, prevent retries; while an intent for the same symbol remains unresolved, prevent submission with a new order ID. & Simulated failures test stale writes, retries, and replacement order IDs. \\
    \bottomrule
  \end{tabular}
\end{table}

\subsection{Portfolio accounting}

Let $w_d\in\mathbb{R}^N$ be the target for risky assets decided after observing the close at decision time $d$, and let $e(d)$ be the first available execution bar strictly after $d$. Holdings are accounting units valued at adjusted closing prices. At $e(d)$, the primary event engine solves for quantities $q_{i,e(d)}$ using net asset value, or NAV, after costs so that the resulting risky allocation matches $w_d$. Between rebalances, quantities stay the same while weights drift as prices and NAV change. Let $P_{i,t}$ be the adjusted closing price and $V_t$ portfolio NAV. The risky weight at the prior close is $h_{i,t-1}=q_{i,t-1}P_{i,t-1}/V_{t-1}$. These holdings first earn returns over the interval following the execution close. With nonnegative risky positions and no borrowing, $h_{i,t}\geq 0$ and $\sum_i h_{i,t}\leq 1$; residual cash is
\begin{equation}
  h^c_t = 1-\sum_{i=1}^N h_{i,t}.
\end{equation}
For simple asset returns $r_{i,t}$ over $(t-1,t]$ and return on the cash account
$r^c_t$, portfolio return before costs is
\begin{equation}
  r^{\mathrm{gross}}_t = \sum_{i=1}^N h_{i,t-1}r_{i,t}
  + h^c_{t-1}r^c_t.
  \label{eq:gross}
\end{equation}
The engine rejects gaps in the cash return series. Let $\Delta a_{i,t}$ be the value of a trade in risky asset $i$ as a fraction of NAV before trading, positive for a purchase and negative for a sale. Define $B_t=\sum_i\max(\Delta a_{i,t},0)$ and $S_t=\sum_i\max(-\Delta a_{i,t},0)$. We measure turnover as $\max(B_t,S_t)$, the larger of purchases and sales. The charge applies to purchases plus sales, $B_t+S_t$, at a rate of 13 basis points: $\kappa_t=0.0013(B_t+S_t)$ of NAV before trading. Let $V^-_t$ be NAV before trading and $V_{t-1}$ NAV at the prior close. The reduction in period return is $d_t=\kappa_t V^-_t/V_{t-1}$ and $r^{\mathrm{net}}_t=r^{\mathrm{gross}}_t-d_t$. We apply the same flat proportional charge to every trade to study sensitivity to trading costs. Execution models balance volatility risk against transaction costs from temporary and permanent market impact \citep{almgren2001execution}.

\begin{samepage}
We report nominal compound annual growth rate, or CAGR, maximum drawdown, annualized turnover, annualized purchases plus sales, and the sample Sharpe ratio relative to benchmark returns $r^f_t$:
\begin{equation}
  \mathrm{SR}_f = \sqrt{252}\,
  \frac{\operatorname{mean}(r_t-r^f_t)}
       {\operatorname{sd}(r_t-r^f_t)}.
  \label{eq:sharpe}
\end{equation}
\end{samepage}
The baseline sets $r^c_t=r^f_t=r^{\mathrm{SHV}}_t$. With zero cash income, $r^c_t=0$; the convention comparisons retain $r^f_t=r^{\mathrm{SHV}}_t$. Thus, the cash comparison changes portfolio income and uses SHV as the benchmark in both runs.

Equation~\ref{eq:sharpe} scales the daily Sharpe ratio by $\sqrt{252}$ to report a descriptive annualized value. Its interpretation as an annual population quantity depends on the return distribution and serial dependence \citep{lo2002statistics}. For uncertainty estimates, we use the circular block bootstrap \citep{politis1992circular}. Sampling consecutive dates together preserves dependence within each block. Sampling blocks with replacement yields a new return history; when a block reaches the sample's end, it continues from the beginning.

Capital left in cash earns the return on SHV in our accounting. SHV is a Treasury ETF, so that return includes fund expenses, duration exposure, and market price changes. We calculate its return from adjusted closing prices and use the same price basis for risky holdings. These accounting units track total return. Executable share purchases depend on the market prices and execution records for the trades.

\subsection{Return decomposition}

We break down net return above cash into allocation, timing, passive exposure, and cost. Let $x_t=\sum_i h_{i,t-1}$ be risky exposure during return period $t$,
let $b_t$ be the return of a buy-and-hold basket with equal initial weights, and let
$\bar{x}$ be the mean of $x_t$ over the full sample. Basket returns at actual exposure and at mean exposure are
\begin{align}
  m_t &= x_t b_t + (1-x_t)r^c_t, \\
  k_t &= \bar{x} b_t + (1-\bar{x})r^c_t.
\end{align}
Then net return above cash has the exact arithmetic decomposition
\begin{align}
  r^{\mathrm{net}}_t-r^c_t
  ={}& \underbrace{(r^{\mathrm{gross}}_t-m_t)}_{\text{active risky allocation}}
  + \underbrace{(m_t-k_t)}_{\text{dynamic exposure timing}} \nonumber\\
  &+ \underbrace{(k_t-r^c_t)}_{\text{passive risky exposure}}
  + \underbrace{(r^{\mathrm{net}}_t-r^{\mathrm{gross}}_t)}_{\text{implementation cost}}.
  \label{eq:decomposition}
\end{align}
The allocation term uses the basket return at the strategy's actual exposure as its reference. It combines group selection, weights within the risky portfolio, and drift between rebalances. Brinson and Fachler's stock selection term weights return differences within groups by benchmark group weights \citep{brinson1985measuring}.

The timing contribution is positive when the strategy holds a larger fraction of capital in risky assets on days when the basket earns more than cash and a smaller fraction on days when it earns less. Exposure is measured relative to its mean over the entire evaluation period. This makes the term a measure of historical association between exposure and returns above cash. Interpreting that association as a causal effect, or turning it into a tradable timing rule, would involve assumptions beyond the accounting.

Multiplying the mean of each daily component by 252 expresses its contribution at an annual rate. The contributions add to the annualized arithmetic mean of return above cash. We use this mean to compare accounting effects; CAGR answers a different question by compounding the sequence of daily returns.

For each resample, we use the same dates for all five series in Equation~\ref{eq:decomposition}, preserving the daily equality and the sum of component means. We then calculate percentile endpoints separately for each component and for net return above cash. The component endpoints generally do not add to the endpoints for the net return because each series has its own ordering across resamples. The proportion of positive resamples counts how often the mean exceeds zero in this distribution.

\subsection{Accounting comparisons}

To isolate accounting effects, we rerun the same portfolio decisions and prices through the primary engine while changing one convention at a time. Cash earns zero in one comparison, and trading carries zero cost in another. A third assigns each new allocation to the return interval ending at its decision close. The decision uses that close, so the portfolio receives a return from before it could trade. This timing error lets us examine how crediting a past return changes the performance statistic.

Using the same target weights preserves the strategy's decisions, but cash income and trading charges change its wealth. Those changes can alter subsequent exposure and trade size, which the reruns capture. We adapt the comparison design of \citet{zhang2026alpha} by reusing the same portfolio decisions while varying one convention at a time. The basket and engine comparisons address broader changes in allocation or accounting and are interpreted separately.

\subsection{Validation checks}

Validation begins with a complete allocation record and checks allocation sums, position limits, exposure after overlays, and information availability. A comparison portfolio follows the same target exposure schedule in each study. Matching the exposure decisions lets us examine how the allocation across assets affects performance.

Synthetic examples compare the accounting with answers calculated by hand. Experiment records identify the inputs, settings, and software environment needed to repeat each calculation. Separate order tests introduce stale state, an uncertain submission outcome, and a replacement order for the same symbol. The system retains the original order request and waits until its outcome is reconciled before submitting another order for the same symbol. Appendix~\ref{app:scope} distinguishes these tests from validation of live execution.

\section{Evaluation design}

\subsection{Retrospective audit}

The first study evaluates a weekly rotation strategy on six ETFs from January 2, 2018, through December 31, 2025. It combines group ranking, residual momentum, exposure states from a Gaussian mixture, and volatility scaling. The strategy predates the audit, and the number of earlier trials is unknown. We retain the strategy as a retrospective case for investigating accounting effects.

QQQ is both an eligible asset and the ranking benchmark. Regressing its returns on themselves gives zero residuals. Rounding error can leave a small positive score that receives substantial weight after normalization. We retain the original portfolio decisions for the audit and evaluate the correction in Appendix~\ref{app:score-correction}. The specification also uses two copies of the same volatility series at different scales, short early fitting windows, and one mixture seed. Appendix~\ref{app:retrospective} gives the rule.

The event engine credits cash with SHV returns, charges 13 basis points on purchases and sales, and executes each weekly allocation at the first later close. The decomposition uses the return of a basket at the strategy's actual exposure before costs. A separate comparison portfolio starts with equal basket weights, lets their composition drift, follows the strategy's target exposure, and pays its own trading charges. Both portfolios use the same execution timing and cash convention. This comparison measures the contribution of allocation choices after costs along the strategy's exposure schedule.

\subsection{Comparison across ETF sets}
\label{sec:portfolio-study}

We wrote the second study's analysis protocol before requesting its data. The market history itself was already available during development. This ordering reduces discretion over the subsequent analysis while leaving the study a historical comparison. Section~\ref{sec:limitations} discusses historical reuse and external registration.

Two ETF sets use adjusted closing prices from January 2, 2014, through December 31, 2025. The sector set contains XLB, XLE, XLF, XLI, XLK, XLP, XLU, XLV, and XLY. The country set contains EWA, EWC, EWG, EWH, EWJ, EWL, EWP, EWQ, EWS, and EWU. Both sets also include SHV. Evaluation covers \ExpansionEvaluationSessions{} sessions from January 2, 2018, through December 31, 2025. Each dataset keeps only sessions with observations for every symbol in that ETF set, including SHV. Preprocessing excludes forward filling and symbol substitution. Both sets consist of surviving funds and share market exposures.

Table~\ref{tab:expansion-design} gives the target rules. Each monthly decision uses the first complete session and executes on the first strictly later close. Risky weights are nonnegative and sum to at most one; idle capital earns SHV returns. The evaluator checks the target weights before calculating returns. The event engine values holdings at adjusted closing prices and changes their quantities only at rebalances. It charges 13 basis points for every dollar bought or sold, using the $B_t+S_t$ cost base defined above.

At each decision, the comparator preserves target gross risky exposure and divides it equally across the risky ETFs, excluding SHV. It uses the same execution timing and cash convention and pays its own trading charges.

\begin{table}[t]
  \caption{Portfolio rules for the sector and country ETF sets. In the plots, TS momentum denotes time series momentum and GBRT denotes regression trees fit by gradient boosting. Ties are resolved by ticker symbol in alphabetical order. Reversal and gradient boosting assign one-third to each qualifying asset; the remainder stays in cash when fewer than three qualify.}
  \label{tab:expansion-design}
  \centering
  \begin{tabular}{>{\raggedright\arraybackslash}p{0.22\linewidth}
                  >{\raggedright\arraybackslash}p{0.70\linewidth}}
    \toprule
    Family & Portfolio rule \\
    \midrule
    Time series momentum & Positive simple returns over 252 sessions; equal weight on qualifying assets; hold cash if none qualify. \\
    Cross-sectional momentum & Three highest simple returns from $t-252$ to $t-21$; one-third each. \\
    Short-term reversal & Up to three lowest negative simple returns over 5 sessions; one-third each, without renormalizing. \\
    Gradient boosting & Up to three highest strictly positive predictions of log return over 21 sessions; one-third each, without renormalizing; five seeds. \\
    \bottomrule
  \end{tabular}
\end{table}

At each monthly decision, we fit one regressor per seed using observations from all risky assets in that ETF set together. A training row is eligible only if its return outcome is observable by that close. Features use prices available by their own decision dates. Appendix~\ref{app:expansion-targets} gives the feature windows, return label, training window, model settings, and seeds.

Each random seed produces a separate portfolio. We calculate performance for each portfolio and then average the five metric values. In particular, the gradient boosting Sharpe entry averages five Sharpe ratios. A portfolio that combines their holdings would have its own return series and Sharpe ratio.

\subsection{Comparison design}

The first three comparisons reuse portfolio decisions while changing timing, cash income, or trading charges separately. Retroactive assignment deliberately credits a new allocation with the return ending at its decision close, introducing a timing error. We subtract baseline performance from each alternative: retroactive assignment, zero cash income, and zero trading charges.

The fourth comparison subtracts the comparison portfolio's performance after its own costs from the net strategy's performance. Both portfolios execute at the first later close and follow the same target exposure, but their allocation across assets, trading activity, and costs differ. The fifth reports vectorized minus event engine performance, combining differences in daily weight resets, weight drift, trade accounting, and sizing after costs. Combinations of the timing, cash, and cost changes are outside this design.

Timing also affects whether the initial entry charge falls inside the evaluation period. When the first target is nonzero, retroactive assignment places that initial purchase before the window. Execution at the first later close places it inside. Both the entry charge and the return timing contribute to the comparison.

We compare annualized mean return and sample Sharpe relative to SHV. We resample daily observations on common dates in circular blocks \citep{politis1992circular}. Pointwise 95\% intervals use the 2.5th and 97.5th percentiles of the metric differences across \ExpansionBootstrapReplications{} resamples. Each primary block contains 21 consecutive sessions; sensitivity checks use 5 and 63. The same dates apply to both return paths and SHV, and to every regression seed, preserving their contemporaneous relationships. We recompute the metrics from each resample without refitting the models or regenerating portfolio decisions. We interpret each interval conditional on the ETF sets, strategy families, design choices, and seeds in the study. Section~\ref{sec:limitations} addresses uncertainty beyond this calculation and inference across multiple comparisons.

\section{Results}

\subsection{Performance across ETF sets}

The investment universe changes the performance comparison. All four families earn higher mean return, compound growth, and Sharpe relative to SHV in sectors than in countries over this period. Country time series momentum earns a positive nominal return and a slightly negative Sharpe relative to SHV. The cash benchmark changes the interpretation of that nominal gain.

The risk comparison between gradient boosting and cross-sectional momentum also reverses across ETF sets. In countries, the regression portfolios have lower mean compound growth, and every seed has lower volatility and a shallower drawdown than the momentum rule. In sectors, every seed has higher volatility and a deeper drawdown. Thus, the same fitting procedure produces different risk profiles in the two investment universes. Appendix Table~\ref{tab:expansion-baseline} and Figure~\ref{fig:expansion-baseline} report the estimates; model entries average metrics across separate seed portfolios.

\subsection{Effects of accounting conventions}

Short-term reversal shows the strongest mean return response to retroactive close assignment. Its selection window includes the decision close, so retroactive assignment credits returns from within the information used to select recent losers. This couples the selection signal with the return credited to the new portfolio. The result is consistent with this mechanism: annualized mean return falls by about seven percentage points in each ETF set, and both primary pointwise intervals lie below zero. Sector gradient boosting is the third timing comparison with an interval below zero in Appendix Figure~\ref{fig:contract-return}.

For time series momentum, retroactive assignment raises mean return in both ETF sets, although both pointwise intervals include zero. The contrast with reversal shows why a lower backtest return can still come from crediting returns before execution. Valid timing follows the order of observation, decision, execution, and return. Appendix~\ref{app:timing-example} demonstrates how the same chronology error can favor either side of a comparison.

Cash affects performance through both the amount left uninvested and the dates on which it earns income. Cross-sectional momentum stays almost fully invested and changes little when cash earns zero. Reversal leaves more capital in cash, especially in countries, and loses more income under that convention. Five of the eight pointwise intervals lie below zero. Keeping SHV as the reporting benchmark in both runs isolates the cash convention and its subsequent effects on wealth and holdings.

Trading activity explains much of the variation in cost effects. Removing charges raises annualized mean return by \ExpansionCostEffectRange{}, and all eight pointwise intervals lie above zero. Country reversal buys and sells a much larger value relative to capital each year than sector time series momentum. Multiplying that trading measure by the proportional charge approximates the direct return reduction. Rerunning the accounting also incorporates the effect of costs on subsequent wealth and trade sizes.

\subsection{Strategy rankings}

A change in accounting can change which strategy ranks higher even though the underlying portfolio decisions are identical. Table~\ref{tab:rank-sensitivity} shows one pairwise swap in each ETF set under each alternative. Retroactive assignment exchanges gradient boosting and time series momentum in sectors, and reversal and time series momentum in countries. Removing costs exchanges reversal and time series momentum in sectors and moves gradient boosting above cross-sectional momentum in countries.

A rank change can reflect a very small difference. Under zero trading cost, sector reversal and time series momentum are about 0.006 apart in Sharpe, enough to change their order while both round to the same value at two decimal places. The table keeps three decimals to make that ordering visible. These ranks describe the sample order. The pointwise intervals in Appendix Figure~\ref{fig:contract-sharpe} concern changes in Sharpe within each strategy under a different accounting convention.

\begin{table}[t]
  \centering
  \caption{Sample Sharpe relative to SHV and rank within each ETF set. Rank one is highest. Bold marks the highest sample Sharpe within each set and convention. Each alternative changes timing or trading cost separately from the baseline. Retroactive assignment credits returns before execution and can move the initial entry charge outside the evaluation window. Zero trading cost retains baseline timing and cash income. Gradient boosting entries average five individual seed Sharpes.}
  \label{tab:rank-sensitivity}
  \setlength{\tabcolsep}{5pt}
  \begin{tabular}{@{}lrrrrrr@{}}
    \toprule
    & \multicolumn{2}{c}{Baseline}
    & \multicolumn{2}{c}{Retroactive assignment}
    & \multicolumn{2}{c}{Zero trading cost} \\
    & & & \multicolumn{2}{c}{\textit{returns before execution}} & & \\
    \cmidrule(lr){2-3}\cmidrule(lr){4-5}\cmidrule(l){6-7}
    Strategy & \shortstack{Sharpe\\ratio} & Rank & \shortstack{Sharpe\\ratio} & Rank & \shortstack{Sharpe\\ratio} & Rank \\
    \midrule
    \ExpansionRankRows
    \bottomrule
  \end{tabular}
\end{table}

\subsection{Sources of return}

The basket comparison asks whether a strategy's allocation across risky assets improves performance after costs along its target exposure schedule. Each comparison portfolio follows the same timing and cash convention and pays its own costs. The retrospective basket begins with equal weights and drifts thereafter; the expansion basket returns to equal weights at every decision. Seven of the eight expansion intervals for the difference in mean return include zero. Country time series momentum is the exception: its mean return falls below the basket return, with a pointwise interval below zero. The basket follows the strategy's own exposure schedule, making this a conditional reference for allocation performance. An independent benchmark would have a separate definition, and causal attribution of skill depends on a design that identifies that effect.

For the retrospective strategy, the cash account is central to understanding performance: the portfolio is entirely in cash during \PaperFullyCash{} of the sample. Even with that income, its compound growth falls below SHV; setting cash income to zero deepens the shortfall. The arithmetic decomposition shows a corresponding shortfall in mean return above cash. Allocation and trading costs outweigh the positive contributions from passive exposure and exposure timing.

The allocation contribution has a pointwise interval below zero. Appendix Table~\ref{tab:decomposition} gives the component estimates and intervals. The component means add to mean return above cash before rounding, linking the accounting explanation directly to the strategy's shortfall relative to cash.

The two allocation comparisons use different exposures and cost treatments. The allocation component compares gross returns at the strategy's actual exposure. The separate basket portfolio follows target exposure and incurs its own trades and costs. Those differences explain why its performance gap differs from the allocation component in Appendix~\ref{app:retrospective-results}.

\subsection{Sensitivity checks}

The five regression portfolios in Appendix Figure~\ref{fig:seed-sensitivity} show how a single model specification can lead to different rankings. Showing each run alongside the mean Sharpe makes the dependence on the seed visible.

In sectors, four regression runs exceed time series momentum on Sharpe relative to SHV, and one exceeds cross-sectional momentum. In countries, all five exceed time series momentum and two exceed cross-sectional momentum. Those two country runs also have higher CAGR than cross-sectional momentum, even though every country run has lower mean return. Mean return, variability, and compounding enter these metrics differently, so a model's position depends on both the seed and the metric. Appendix~\ref{app:expansion-followups} examines the corresponding accounting effects, annual returns above cash, and trading activity.

The engine comparison holds the portfolio decisions and prices common and changes how decisions become daily holdings. The vectorized engine calculates each day's return using the last execution weights. The event engine holds asset quantities between trades, allowing weights to drift with prices. Trade accounting and sizing after costs also differ. These are differences in accounting conventions, whose consequences we compare through daily returns and final wealth in Appendix Figure~\ref{fig:engine-sensitivity}. Reference calculations with known answers test individual accounting cases. The engine comparison measures differences between the two conventions; a broader assessment of implementation errors would examine other systems and cases.

The largest daily difference and the largest final wealth difference occur in different cases. Sector gradient boosting has the largest absolute daily return difference, whereas sector cross-sectional momentum has the largest final wealth gap, \ExpansionEngineMaxWealthGap{} of initial capital. Final wealth compounds each engine's return series separately. Daily differences can reinforce or offset one another over time, making the full return path central to the wealth comparison.

The retrospective strategy produces much closer returns across engines. Agreement in that case and larger differences in the expansion study show that the effect of an accounting convention depends on the portfolio decisions it processes.

Changing the resampling block length leaves the timing, cost, and basket intervals in the same categories: below zero, including zero, or above zero. Appendix Table~\ref{tab:effect-counts} reports these categories for blocks of 5, 21, and 63 sessions. Cash and engine comparisons are more sensitive. The small sector gradient boosting cash interval includes zero with the longest blocks; for engine differences in mean return, two intervals lie above zero with blocks of 5 sessions and four with either longer choice.

\subsection{Performance across subperiods}

The retrospective strategy also changes its ordering across metrics over time. It has higher CAGR but lower Sharpe relative to SHV in 2022 through 2025 than in 2018 through 2021. The basket outperforms it on both metrics in each half and over the full period. These subperiod comparisons use the original continuous portfolio path, so they describe when performance occurred within the same strategy history.

\section{Limitations}
\label{sec:limitations}

The studies establish accounting effects for the strategies and histories examined here. Both market histories were available during development. Records of earlier experiments on the rotation strategy are incomplete. We wrote the second protocol before data retrieval but did not register it externally. Evidence from an independent sample would come from a period kept outside development or from future observations. This distinction determines how far the historical comparisons can support a claim about future performance.

The ETF sets share dates and market exposures, and the five seeds repeat one model specification. Our intervals describe variation from resampling dates in the existing portfolio paths. They condition on the models and decisions, leaving model selection and a wider population of seeds outside their scope. Nominal coverage rests on stationarity, sufficiently weak dependence, and suitable moment conditions; simultaneous coverage also involves multiplicity. An adjustment for all prior strategy trials using the Deflated Sharpe Ratio or White Reality Check depends on trial records that are incomplete here \citep{white2000reality,bailey2014deflated}. Appendix~\ref{app:scope} gives the assumptions and their implications.

The audit concerns historical accounting and specific order failures introduced in tests. Live execution adds liquidity, order size, market impact, and venue behavior to the assessment of capacity and execution quality. Those effects extend beyond the flat trading charge used here and call for market and broker observations. Using the historical comparisons as deployment evidence could expose investors to losses through underestimated risk or costs. Their value is in making the assumptions behind a performance claim available for scrutiny.

\section{Conclusion}

A financial model's evaluation depends on the path from a decision to the returns assigned to it. In these studies, a timing error can favor or penalize a strategy, while cash income and trading costs alter both performance and relative rankings. The retrospective decomposition explains a shortfall relative to cash through allocation and trading costs that outweigh gains from exposure. Together, the comparisons show why model assessment begins with the chronology and accounting of the portfolio. Tracing decisions into holdings and cash flows makes the economic meaning of a performance estimate clearer.

\clearpage
\begin{ack}
Both authors contributed funding to this work. We obtained price data from Yahoo Finance through yfinance and thank the yfinance, NumPy, pandas, scikit-learn, and Matplotlib contributors for their software.
Parts of the work were assessed and supported under a contract with micro1. We used the Autonomous Research and Engineering System (ARES) agentic platform and the ARES CLI, both developed by coauthor Kevin Lee, for parts of the development and verification.
We have no competing interests to declare.
\end{ack}

\FloatBarrier
\bibliographystyle{plainnat}
\bibliography{references}
\clearpage

\appendix
\raggedbottom

\section{Scope of the evidence}
\label{app:scope}

Accounting for strategy search starts with the history of candidate strategies. The Deflated Sharpe Ratio uses the effective number of independent trials and the variance across their Sharpe estimates, together with the selected return series' Sharpe ratio, length, skewness, and kurtosis. White's Reality Check uses a universe of candidate strategies, benchmark, performance criterion, and observations that can be resampled together or equivalent summaries \citep{white2000reality,bailey2014deflated}. With an incomplete trial history, we report sample metrics and intervals conditional on the strategies in this study.

The allocation component can also be written as
$A_t=\sum_i h_{i,t-1}(r_{i,t}-b_t)$, a contribution per unit of portfolio capital.
This expression remains defined when the portfolio holds only cash.
For the timing component, let $T_t=m_t-k_t$, $z_t=b_t-r_t^c$, and let
$\bar{z}$ denote the mean over the $n$ evaluation sessions. Then
\begin{equation}
  252\bar{T}
  = \frac{252}{n}\sum_{t=1}^{n}(x_t-\bar{x})(z_t-\bar{z}).
  \label{eq:timing-covariance}
\end{equation}
The mean timing contribution is positive when exposure and basket excess return
move together over the sample. Its annualized mean equals 252 times their
empirical covariance, calculated with denominator $n$. Mean exposure comes from the full evaluation period, so this is a description
of historical association. Causal attribution and a tradable timing strategy
would involve additional assumptions. We compute the daily components once,
then resample them together using the original $\bar{x}$ in every resample.
All component comparisons share that reference exposure.

The decomposition adds daily return contributions and their means. Extending it to compound returns involves a geometric linking rule. Similarly, the proportion of resamples with a positive mean is a frequency in the bootstrap distribution. A posterior probability would be defined within a Bayesian model, and inference across multiple comparisons would account for their multiplicity.

The datasets contain funds available at the time of collection; point-in-time constituent histories would change the investment opportunity set. The country ETFs trade in the U.S. and carry currency exposure as well as overlapping market exposures.

Using SHV for cash income introduces duration exposure, fund expenses, and market price variation. Adjusted closing prices express holdings as units of total return. Translating the same decisions into share quantities and assessing spreads, market impact, volume constraints, and fill uncertainty would use additional market and broker records.

We use one regression specification and five integer seeds, pooling asset observations within each ETF set without a symbol feature. The date resampling starts from the resulting models and portfolio paths. Its nominal coverage rests on stationarity, sufficiently weak dependence, suitable moment conditions, and the block construction. Comparing three block lengths tests sensitivity to the resampling construction. Stationarity and dependence remain assumptions about the returns. The intervals describe variation within the existing paths, conditional on model selection, hyperparameters, and seeds. Inference about future performance would also address how well those paths represent future conditions.

The checks test specific invariants against known answers and specific failures against their expected handling. A defect detection rate would describe performance over a defined collection of faulty implementations. The incremental benefit of a check would come from a comparison with a procedure that omits it. Here the accounting and overlay comparisons measure changes in portfolio returns and exposure.

Order tests introduce stale state, retries after an uncertain outcome, and replacement identifiers. The ledger keeps a later order for the same symbol pending until the earlier intent is reconciled, including when the later order is otherwise valid. Assessing a live system would also involve tests of pending exposure across the account, delivery among concurrent clients, authenticated transport, venue behavior, and deployment controls. The present evidence covers simulated failures in order state.

\subsection{Timing error with two assets}
\label{app:timing-example}

Consider two assets observed at four successive closes, with prices given below. Starting wealth is $1$, held entirely in cash, and both cash return and trading charge are zero. After observing $t_1$, a winner rule selects A, while a loser rule selects B; either rule targets full investment in the selected asset. These rules illustrate one interval. The empirical time series momentum and reversal rules use 252 and 5 sessions, respectively.

\begin{center}
  \begin{tabular}{@{}lrrl@{}}
    \toprule
    Close & Price of A & Price of B & Baseline chronology \\
    \midrule
    $t_0$ & 100 & 100 & Entirely in cash \\
    $t_1$ & 110 & 90 & Observe prices and select A or B \\
    $t_2$ & 110 & 90 & Execute the selected target \\
    $t_3$ & 110 & 90 & First return earned by the new holding \\
    \bottomrule
  \end{tabular}
\end{center}

Under the baseline, execution occurs at $t_2$. The new holding first earns the return ending at $t_3$, which is $110/110-1=0$ for A and $90/90-1=0$ for B. Both portfolios finish with wealth $1.00$.

Retroactive assignment instead credits the return over the interval from $t_0$ to $t_1$, which has already ended: $110/100-1=+0.10$ for A or $90/100-1=-0.10$ for B. Subsequent prices are unchanged, so the fictitious terminal wealth is $1.10$ or $0.90$.

The same chronology error creates a gain for the winner rule and a loss for the loser rule. This example isolates return timing. The empirical comparison uses a trading charge of 13 basis points and includes any shift of the initial entry charge across the evaluation boundary.

\section{Retrospective case specification}
\label{app:retrospective}

The universe is QQQ, VGT, GLD, TLT, SPY, and VIG, grouped as growth equities, gold and nominal duration, and broad and dividend equities. On the first session of each week, the strategy keeps the top two groups by an information ratio over 126 sessions relative to QQQ. Within selected groups, it estimates a regression on one factor over 126 sessions, then sums residual returns over the next 126 sessions. The most recent 21 sessions are excluded from score formation. Each residual equals the asset return minus the intercept and benchmark contribution from that regression. Residuals enter the score in return units with coefficient one. Positive scores receive nonnegative weights in proportion to their scores, capped at 60\% per asset.

A Gaussian mixture with three components and full covariance uses seed 42, covariance regularization $10^{-5}$, and at most 100 iterations. It maps QQQ return and the same volatility series over 20 sessions at two scales to zero, half, or full exposure. A separate scaler uses QQQ's annualized realized volatility over 21 sessions, expressed as a percentage. It divides 20 by this value and clips the exposure multiplier to $[0.3,1.0]$.

Evaluation begins after at least 274 sessions of history; this sample provides 503. When a benchmark, group signal, or regression estimate is unavailable, the strategy falls back to plain momentum. It holds cash when all mature residual scores in those groups are nonpositive. The overlay comparisons remove the regime rule, the volatility scaler, or both.

\clearpage
\subsection{Retrospective accounting results}
\label{app:retrospective-results}

\begin{table}[h]
  \caption{Retrospective performance from 2018 through 2025. All Sharpe ratios in this table use SHV as the reporting benchmark, including the strategy with zero return on residual cash. That strategy's Sharpe against zero is \PaperZeroCashSharpe{}. The comparison portfolio executes at the first later close and matches the strategy's target exposure.}
  \centering
  \setlength{\tabcolsep}{4pt}
  \begin{tabular}{@{}lrrr@{}}
    \toprule
    Series & \shortstack{Compound\\annual\\growth rate} & \shortstack{Sharpe\\ratio} & \shortstack{Maximum\\drawdown} \\
    \midrule
    Strategy, after costs, with SHV cash & \PaperNetCAGR & \PaperNetSharpe & \PaperNetMaxDrawdown \\
    Strategy, before costs, with SHV cash & \PaperGrossCAGR & \PaperGrossSharpe & \PaperGrossMaxDrawdown \\
    SHV Treasury ETF & \PaperCashCAGR & Undefined & \PaperCashMaxDrawdown \\
    Basket at actual exposure, before costs & \PaperMatchedCAGR & \PaperMatchedSharpe & \PaperMatchedMaxDrawdown \\
    Comparison portfolio, after costs & \PaperCostedComparatorCAGR & \PaperCostedComparatorSharpe & \PaperCostedComparatorMaxDrawdown \\
    Strategy, after costs, with zero cash return & \PaperZeroCashCAGR & \PaperZeroCashSHVSharpe & \PaperZeroCashMaxDrawdown \\
    \bottomrule
  \end{tabular}
\end{table}

The strategy has a lower mean return than the comparison portfolio after each pays its own costs. The mean daily return difference, multiplied by 252, is \PaperCostedActiveMean{}, with a pointwise 95\% percentile interval from \PaperCostedActiveLower{} to \PaperCostedActiveUpper{}. The interval uses circular blocks of 21 sessions from the original portfolio paths.

\begin{table}[h]
  \caption{Decomposition of mean net return above SHV, annualized by multiplying daily means by 252. Pointwise intervals use the same circular blocks of 21 sessions for every component. The daily equality holds in every resample, and the component means add to mean net return above SHV before rounding.}
  \label{tab:decomposition}
  \centering
  \begin{tabular}{lrrr}
    \toprule
    Component & Mean & 95\% interval & \shortstack{Resamples with\\a positive mean} \\
    \midrule
    \PaperDecompositionRows
    \bottomrule
  \end{tabular}
\end{table}

\begin{center}
  \centering
  \leavevmode\vbox{\refstepcounter{figure}%
  \hbox{\includegraphics[width=0.92\textwidth]{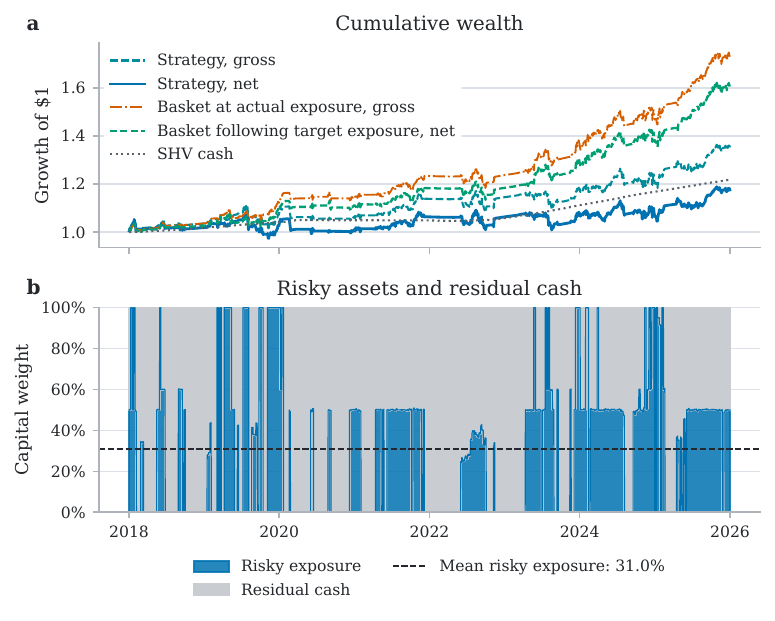}}}
  \parbox{0.92\textwidth}{\small Figure~\thefigure: Growth of one dollar and allocation to risky assets and residual cash in the retrospective study. Exposure is indexed by the ending date of each return interval. Cash earns SHV returns. Residual cash accounts for most capital on average.}
\end{center}

\begin{center}
  \centering
  \leavevmode\vbox{\refstepcounter{figure}%
  \hbox{\includegraphics[width=0.92\textwidth]{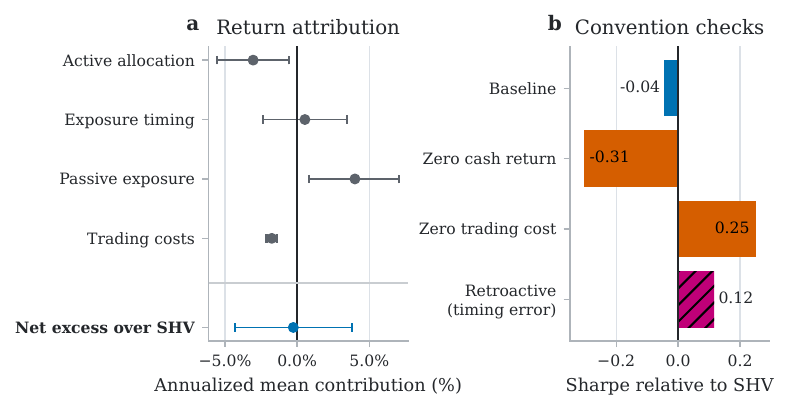}}}
  \parbox{0.92\textwidth}{\small Figure~\thefigure: Retrospective return decomposition and accounting comparisons. Subplot a shows contributions to mean return above SHV at an annual rate. The pointwise 95\% intervals use the same circular resamples with blocks of 21 sessions. Mean net return above SHV equals the sum of the four component means and has its own interval. Subplot b compares Sharpe ratios relative to SHV. Diagonal stripes mark the timing error from applying a new allocation to the return ending at its decision close.}
\end{center}

\subsection{QQQ score correction}
\label{app:score-correction}

For QQQ, the asset and benchmark returns are identical. With positive variance in the estimation window, the regression has slope one and intercept zero, so every residual is zero. Numerical rounding can leave a small positive residual momentum score. The allocation rule divides positive scores by their sum, allowing this rounding error to give QQQ a substantial weight.

Whenever an asset's prices match the benchmark exactly and the regression is valid, we set its residual momentum score to zero. Scores for other assets and fallbacks for unavailable estimates stay unchanged. We rerun the full strategy on the same prices and dates with all other model and accounting settings unchanged. The original portfolio decisions remain the basis of the historical accounting comparisons.

\begin{table}[h]
  \centering
  \caption{Effect of correcting the QQQ score over the same historical sample. Growth and Sharpe use returns after costs; Sharpe is relative to SHV. Mean risky exposure uses holdings entering each return interval.}
  \label{tab:score-correction}
  \begin{tabular}{@{}lrrr@{}}
    \toprule
    Version & \shortstack{Compound annual\\growth} & \shortstack{Sharpe\\ratio} & \shortstack{Mean risky\\exposure} \\
    \midrule
    \PaperScoreCorrectionRows
    \bottomrule
  \end{tabular}
\end{table}

Correcting QQQ's score reduces risky exposure and turnover, while compound growth and Sharpe rise modestly over this sample. Sharpe remains close to zero, and maximum drawdown is slightly deeper.

\section{Code and data availability}
\label{app:reproducibility}

We provide the implementation, aggregate results, software environment, and synthetic examples for examining the accounting. Replicating the empirical returns also depends on the original prices and portfolio decisions, which we do not distribute. Prices came from Yahoo Finance through yfinance; later downloads may include revisions to historical values. We have not confirmed provider permission to redistribute the prices or publish the aggregates.

\section{Expansion results}

\begin{center}
  \parbox{\textwidth}{\refstepcounter{table}\label{tab:expansion-baseline}%
  \small Table~\thetable: Expansion performance after the proportional trading charge, with Sharpe ratios relative to SHV. Bold marks the highest mean return, Sharpe ratio, and compound growth and the shallowest drawdown within each ETF set, using unrounded values. These ranks describe the sample; the subsequent comparisons assess uncertainty in accounting effects. For gradient boosting, we calculate each metric separately for each of the five seeds, then average the five values; this also applies to compound annual growth and maximum drawdown.}
  \smallskip
  \setlength{\tabcolsep}{3.5pt}
  \begin{tabular}{@{}llrrrr@{}}
    \toprule
    Fund set & Strategy & \shortstack{Annualized\\mean return} & \shortstack{Sharpe\\ratio} & \shortstack{Compound\\annual\\growth rate} & \shortstack{Maximum\\drawdown} \\
    \midrule
    \ExpansionBaselineRows
    \bottomrule
  \end{tabular}
\end{center}

\begin{center}
  \parbox{\textwidth}{\refstepcounter{table}\label{tab:effect-counts}%
  \small Table~\thetable: Sensitivity to resampling block length. Each cell lists three counts of pointwise intervals, in order: below zero, including zero, and above zero. These counts cover eight dependent combinations of strategy family and ETF set. Return differences compare annualized means; Sharpe differences use SHV as the benchmark. Only the number of consecutive sessions per resampling block changes. The same models and portfolio paths underlie every column. The nominal coverage applies to each comparison separately, without adjustment for simultaneous coverage.}
  \smallskip
  \begin{tabular}{@{}lccc@{}}
    \toprule
    Metric & 5 sessions & 21 sessions & 63 sessions \\
    \midrule
    \ExpansionEffectRows
    \bottomrule
  \end{tabular}
\end{center}

A return improvement can come with a lower Sharpe ratio when variability changes. With the primary blocks of 21 sessions, the return and Sharpe intervals in Figures~\ref{fig:contract-return} and~\ref{fig:contract-sharpe} have the same classification in all 40 comparisons: below zero, including zero, or above zero. The point estimates share a sign in 39 comparisons. The exception is the sector regression model against its comparison portfolios after trading costs: its annualized mean return is higher by 1.04 percentage points, but its mean Sharpe is lower by 0.023. Both intervals include zero.

For two return paths on the same dates, let $\mu_j=252\,\operatorname{mean}(r_j-r^{\mathrm{SHV}})$
and $\sigma_j=\sqrt{252}\,\operatorname{sd}(r_j-r^{\mathrm{SHV}})>0$
for $j\in\{A,B\}$, using the sample standard deviation. Then
\begin{equation}
  \mathrm{SR}_A-\mathrm{SR}_B
  = \frac{\mu_A-\mu_B}{\sigma_A}
  + \mu_B\left(\frac{1}{\sigma_A}-\frac{1}{\sigma_B}\right).
  \label{eq:paired-sharpe}
\end{equation}
When both series of returns above SHV have the same standard deviation, the Sharpe difference and the mean return difference have the same sign. Unequal standard deviations can reverse the ordering. Since both series subtract the same SHV return, $\mu_A-\mu_B$ is also the difference in annualized portfolio mean returns. For gradient boosting, this equality holds within each seed pair before the differences are averaged.

\begin{center}
  \centering
  \leavevmode\vbox{\refstepcounter{figure}%
  \label{fig:expansion-baseline}%
  \hbox{\includegraphics[width=0.90\textwidth]{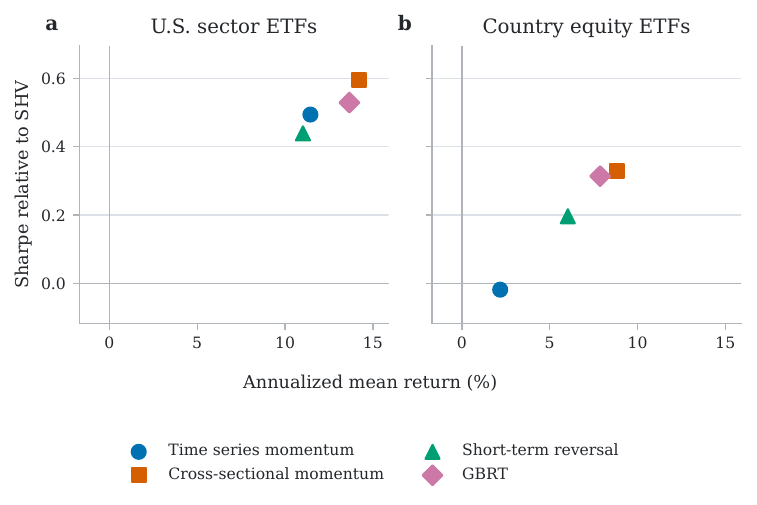}}}
  \parbox{0.90\textwidth}{\small Figure~\thefigure: Annualized mean return and Sharpe relative to SHV after trading costs for the eight expansion cases. Returns are nominal. Each coordinate for gradient boosting averages that metric across five seeds, as in Table~\ref{tab:expansion-baseline}. Figure~\ref{fig:seed-sensitivity} shows the individual seed Sharpes alongside the three rules.}
\end{center}

\begin{center}
  \centering
  \leavevmode\vbox{\refstepcounter{figure}%
  \label{fig:contract-return}%
  \hbox{\includegraphics[width=\textwidth]{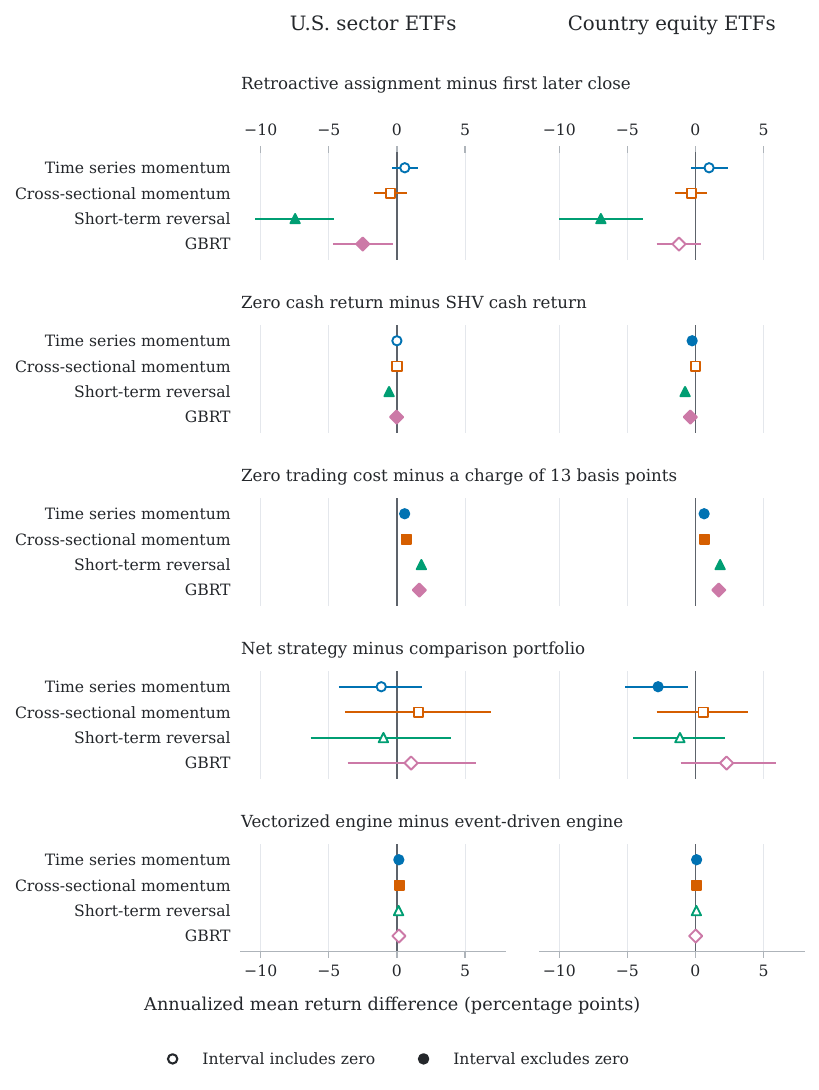}}}
  \parbox{\textwidth}{\small Figure~\thefigure: Changes in annualized mean return, in percentage points, using the subtraction order in each group heading. Lines show pointwise 95\% percentile intervals from common circular resamples with blocks of 21 sessions. The resampling uses the original models, portfolio paths, and seeds. Gradient boosting points average five differences, one per seed. Hollow markers denote intervals that include zero before rounding; some intervals are narrower than their markers. Retroactive assignment credits returns from before execution and can place the initial entry charge outside the evaluation window. SHV is the reporting benchmark in both cash conventions.}
\end{center}

\begin{center}
  \centering
  \leavevmode\vbox{\refstepcounter{figure}%
  \label{fig:contract-sharpe}%
  \hbox{\includegraphics[width=\textwidth]{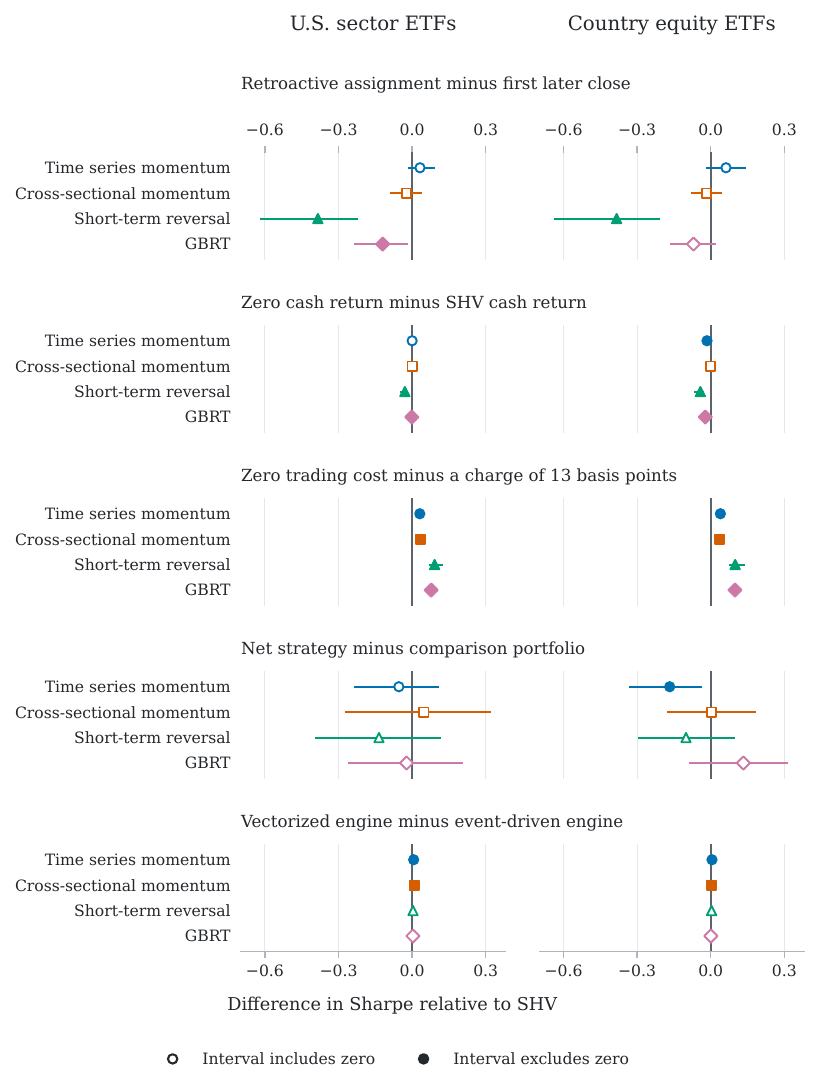}}}
  \parbox{\textwidth}{\small Figure~\thefigure: Changes in Sharpe relative to SHV under the five comparisons. Lines show pointwise 95\% percentile intervals from \ExpansionBootstrapReplications{} circular resamples with blocks of 21 sessions. Each resample uses the same dates for both return paths, SHV, and all five seed portfolios, without refitting models or changing their portfolio decisions. Gradient boosting points average the five individual Sharpe differences. Hollow markers denote intervals that include zero before rounding; some intervals are narrower than their markers. Retroactive assignment credits the portfolio with returns from before execution. Changing residual cash income leaves SHV as the Sharpe benchmark.}
\end{center}

\begin{center}
  \centering
  \leavevmode\vbox{\refstepcounter{figure}%
  \label{fig:seed-sensitivity}%
  \hbox{\includegraphics[width=0.90\textwidth]{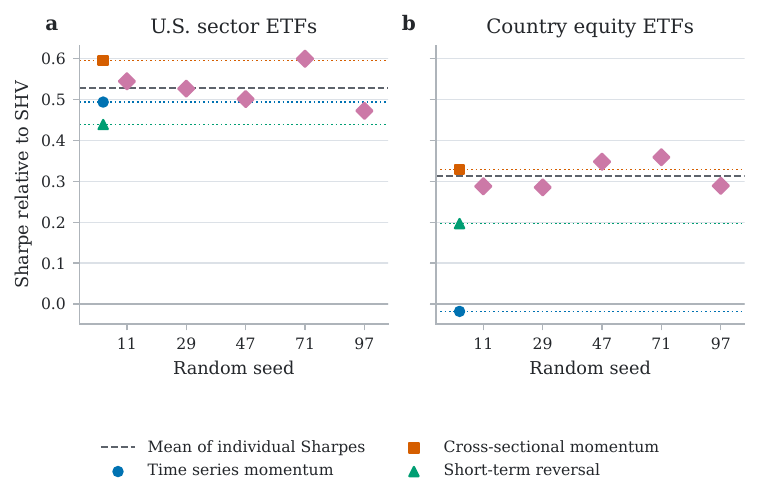}}}
  \parbox{0.90\textwidth}{\small Figure~\thefigure: Baseline Sharpe relative to SHV for five model seeds in each ETF set. Diamonds show the regression model's Sharpe for each seed; dashed lines show the mean of the five Sharpes. Dotted lines and their leftmost markers show each rule's Sharpe for comparison with the seed values. The spread describes these five model runs on the same market history.}
\end{center}

\begin{center}
  \centering
  \leavevmode\vbox{\refstepcounter{figure}%
  \label{fig:engine-sensitivity}%
  \hbox{\includegraphics[width=\textwidth]{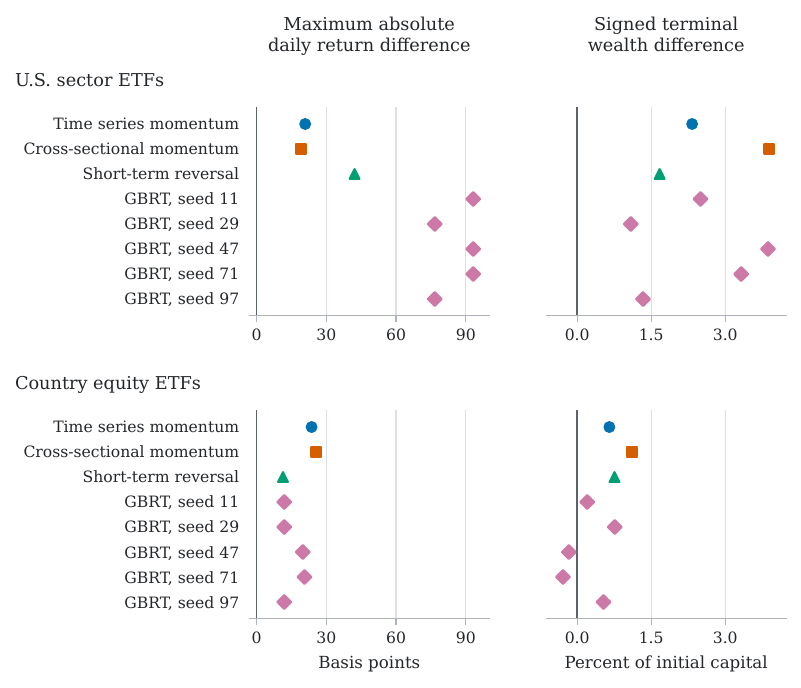}}}
  \parbox{\textwidth}{\small Figure~\thefigure: Engine differences for each rule and each seed of the regression model. The largest absolute daily return difference is in basis points. Final wealth differences are calculated as vectorized engine wealth minus event engine wealth and expressed as a percentage of initial capital. A large daily difference can coexist with a small final wealth difference as daily return differences partly offset over time. The event engine holds asset quantities between trades, allowing weights to drift; the vectorized engine calculates each day's return using the last execution weights. Trade accounting and sizing after costs also differ.}
\end{center}

\section{Retrospective block length sensitivity}

\begin{center}
  \centering
  \leavevmode\vbox{\refstepcounter{figure}%
  \label{fig:block-sensitivity}%
  \hbox{\includegraphics[width=0.88\textwidth]{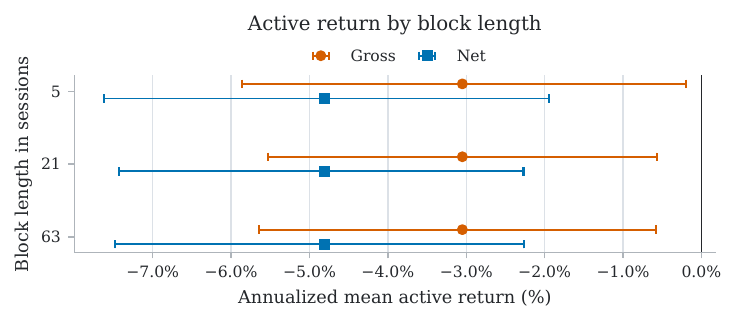}}}
  \parbox{0.88\textwidth}{\small Figure~\thefigure: Annualized mean active return relative to the basket before costs at the strategy's actual exposure. Intervals are pointwise 95\% bootstrap percentiles for circular blocks of 5, 21, and 63 sessions. Gross is strategy return minus basket return, both before trading costs. Net is Gross minus the strategy's trading costs.}
\end{center}

\section{Retrospective engine comparison}

\begin{center}
  \centering
  \leavevmode\vbox{\refstepcounter{figure}%
  \label{fig:engine-comparison}%
  \hbox{\includegraphics[width=0.92\textwidth]{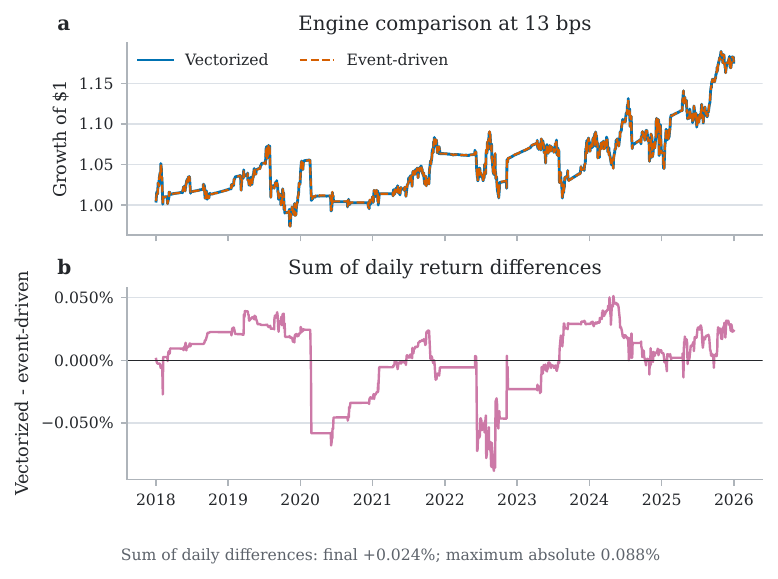}}}
  \parbox{0.92\textwidth}{\small Figure~\thefigure: Subplot a shows growth of one dollar under the two engines. Subplot b adds the daily return differences through each date, subtracting event engine returns from vectorized engine returns. Subplot a compounds each return series separately, so the two subplots use different aggregation rules. The event engine values holdings at adjusted closing prices and leaves their quantities unchanged between rebalances, allowing weights to drift.}
\end{center}

\clearpage
\section{Retrospective cost and overlay sensitivity}

\begin{center}
  \centering
  \leavevmode\vbox{\refstepcounter{figure}%
  \label{fig:sensitivity}%
  \hbox{\includegraphics[width=0.92\textwidth]{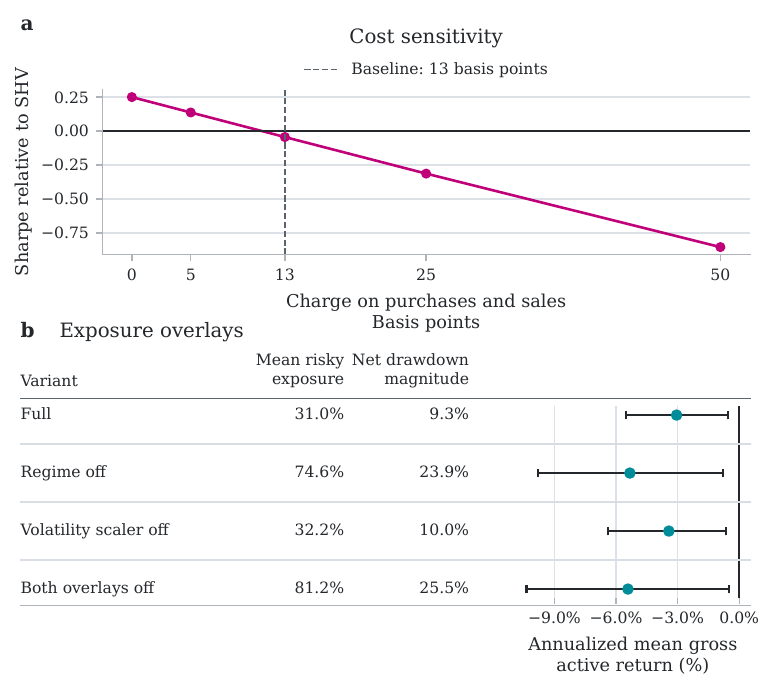}}}
  \parbox{0.92\textwidth}{\small Figure~\thefigure: Retrospective sensitivity to trading costs and exposure overlays. Subplot a gives Sharpe relative to SHV under proportional charges on purchases and sales. For each variant, subplot b shows gross annualized mean active return relative to the basket before costs at that variant's actual exposure. Pointwise 95\% intervals estimate each variant's mean active return relative to its own basket, using the same circular resamples with blocks of 21 sessions. The columns give mean risky exposure and maximum net drawdown magnitude. An interval for the effect of an overlay would compare the variants directly.}
\end{center}

The overlay comparisons retain the original strategy and its signal defect. Removing the regime rule raises mean exposure and deepens maximum drawdown, with larger changes when both overlays are removed. All four intervals for gross active return remain below zero relative to the basket at each variant's actual exposure. Changing the overlays alters both exposure and the return path.

The cost grid illustrates how nominal growth and performance relative to cash can diverge. Sharpe relative to SHV moves from 0.14 at a charge of 5 basis points to $-0.04$ at 13 basis points per dollar bought or sold. CAGR stays positive at 0.40\% with a charge of 25 basis points and falls to $-2.93\%$ at 50. Each point comes from a complete accounting rerun of the same target weights, so the comparison includes the subsequent wealth effects of all five proportional charges.

\clearpage
\section{Portfolio rules across ETF sets}
\label{app:expansion-targets}

Time series momentum keeps every asset with a positive simple return over 252 sessions and assigns equal weight among those assets. If none qualify, the target is cash. Cross-sectional momentum ranks simple returns from session $t-252$ to session $t-21$, breaks ties by ascending symbol, and assigns one-third to each of the three assets with the highest returns. At each decision, the rule targets three assets, including any with negative returns over the ranking window if they are in the top three.

Short-term reversal ranks simple returns over 5 sessions and keeps up to three assets with the lowest strictly negative returns. Each receives one-third of capital; the remainder stays in cash.

A single gradient boosting specification produces the regression model's target weights through the same allocation record interface as the three rules. Each allocation record includes the decision time, symbol, and target weight.

For each decision, we calculate and store features from prices available at that close. The features comprise log returns over 5, 21, 63, 126, and 252 sessions, the sample standard deviation of daily log returns over 21 and 63 sessions, the price divided by its high over the trailing 252 sessions minus one, and ordinal cross-sectional ranks of log returns over 21 and 252 sessions. The ranks use the scale $[0,1]$. The model uses these features without symbol indicators, standardization, or winsorization. The label is the log return over the following 21 sessions. It starts at the decision close, one close before portfolio entry, so the forecast window differs from the executable holding period.

At each monthly decision, we fit a separate model for each seed to asset observations pooled within the ETF set. Only rows whose label endpoint falls on or before that decision close are eligible for training. We use up to the latest 60 decision months with mature labels and require at least 24. Each fit runs under a limit of one thread and uses 100 trees of depth 2, a learning rate of 0.03, a minimum of 10 observations per leaf, a subsampling fraction of 0.8, and squared error loss. The seeds are 11, 29, 47, 71, and 97.

After fitting, the model selects up to three assets with the highest strictly positive predictions, breaks ties by ascending symbol, and assigns one-third of capital to each. The remainder stays in cash. To check information timing, a perturbation test changes prices after a cutoff and verifies that the target at the cutoff is unchanged. The result checks information timing for this implementation at the cutoff in the test. Other implementations and future decisions depend on the availability of their own inputs at decision time.

\clearpage
\section{Historical variation across ETF portfolios}
\label{app:expansion-followups}

We added these descriptive analyses during revision after inspecting the aggregate results. They follow the original portfolio paths under the same ETF sets, model settings, and accounting rules. Calendar years show when performance occurred, while the five seeds show variation among model runs, one source of variability in machine learning benchmarks \citep{bouthillier2021variance}.

\subsection{Performance by calendar year}

For each baseline run and year $y$, we calculate
\begin{equation}
  A_y = \frac{252}{n_y}\sum_{t\in y}
  \left(r_t^{\mathrm{net}}-r_t^{\mathrm{SHV}}\right),
  \label{eq:annual-excess}
\end{equation}
where $n_y$ counts the observed sessions in that year. Dividing the history by year preserves its holdings, costs, and returns across year ends.
Weighting the annual values by their session counts recovers the full sample
annualized mean return above SHV, $\sum_y n_y A_y / \sum_y n_y$. Figure~\ref{fig:yearly-performance} shows every calendar year
in the evaluation window. Entries for the regression model average the five individual seed values.

The ordering over the full sample combines different annual outcomes on the same history. All four country strategies have negative annualized excess returns over SHV in 2018 and 2022, and positive values in 2019, 2021, 2023, and 2025. The sector strategies diverge in 2022: cross-sectional momentum and reversal earn positive mean returns above SHV, while the mean for the regression model is negative. Gradient boosting leads the sector set in 2021 and 2024; cross-sectional momentum leads in 2020, 2022, and 2025.

\begin{center}
  \leavevmode\vbox{\refstepcounter{figure}%
  \label{fig:yearly-performance}%
  \hbox{\includegraphics[width=\textwidth]{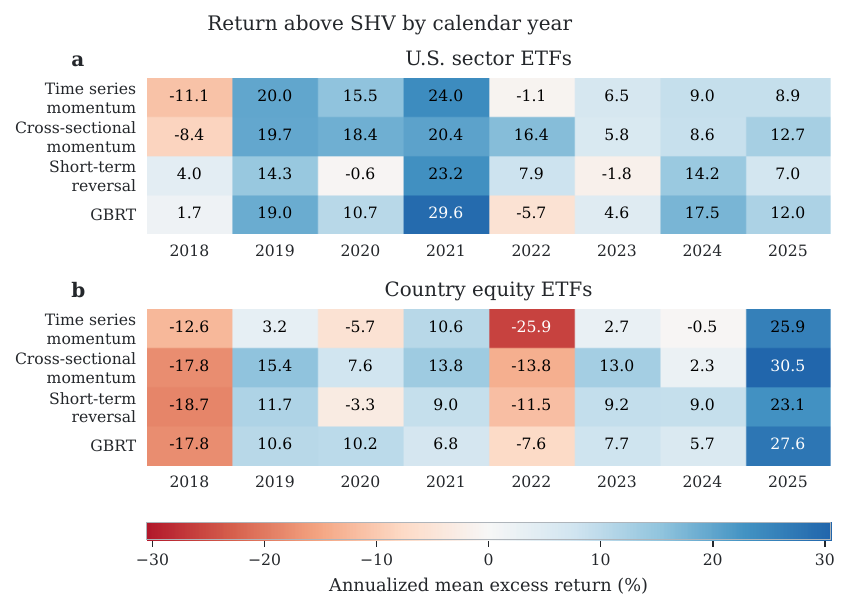}}}
  \parbox{\textwidth}{\small Figure~\thefigure: Baseline performance by calendar year as defined in Equation~\ref{eq:annual-excess}, expressed in percent per year. Each cell is the annualized mean of daily portfolio returns after costs minus SHV returns on the same dates, from the original continuous portfolio path. Cells for the regression model average five individual seed values. Both subplots use the same diverging color scale, with zero at its midpoint. The years describe variation within the historical evaluation sample.}
\end{center}

\clearpage
\subsection{Accounting effects by seed}

Figure~\ref{fig:seed-contract-effects} separates the accounting comparisons by seed. Each cell is a Sharpe difference for one seed and ETF set under the subtraction order for that column. Averaging the five values in a column recovers the regression model's point estimate in Figure~\ref{fig:contract-sharpe}. This view shows whether that average reflects a consistent response across model runs.

The seed average can conceal opposite responses. Against the comparison portfolio after costs, seed 71 has a positive Sharpe difference in sectors and the other four have negative differences. Every country seed has a positive difference. Country engine differences also have both signs despite a positive mean. Cash, cost, and retroactive timing effects are more consistent: each has the same sign across all ten regression paths.

Each row follows monthly fits with one integer seed. All five runs use the same market observations and training specification, so their spread describes variation among those portfolio paths. The intervals in Figure~\ref{fig:contract-sharpe} answer a separate question by resampling dates within those paths. Together, the two views distinguish variation among model runs from uncertainty in an accounting comparison on a given history.

\begin{center}
  \leavevmode\vbox{\refstepcounter{figure}%
  \label{fig:seed-contract-effects}%
  \hbox{\includegraphics[width=\textwidth]{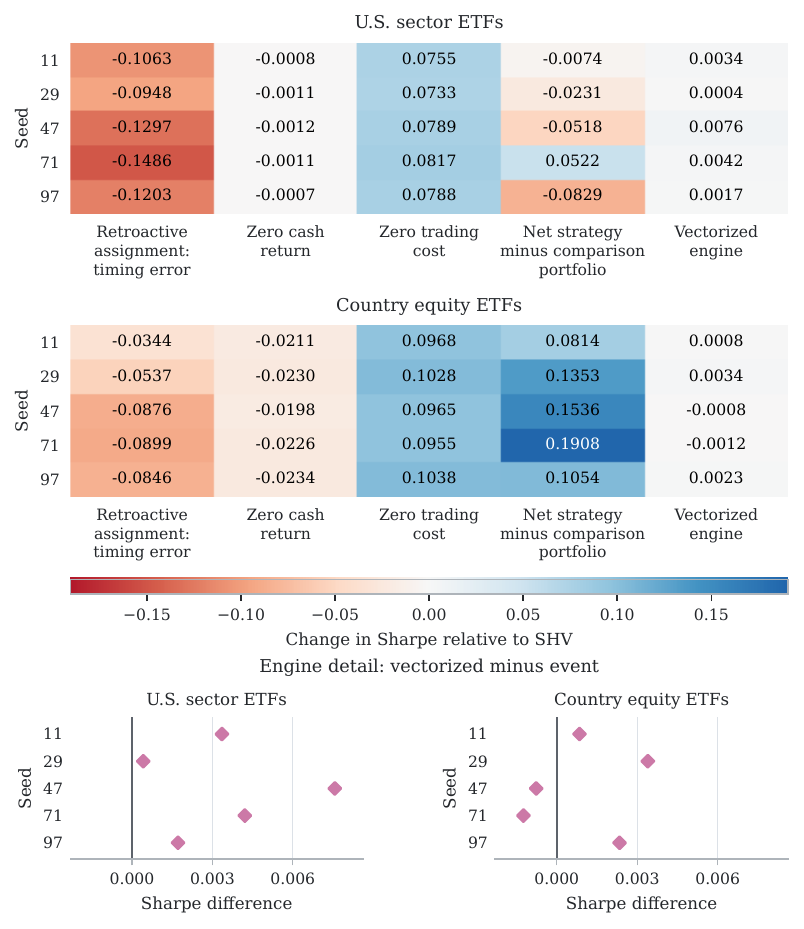}}}
  \parbox{\textwidth}{\small Figure~\thefigure: Changes in Sharpe relative to SHV within each of the five seeds for the regression model. Each matrix column subtracts baseline Sharpe from that convention's Sharpe, except the comparison portfolio column, which reports strategy minus comparison portfolio, both after trading costs. The comparison portfolio matches that seed's target exposure. Retroactive assignment credits returns before execution and may also move the initial entry charge outside the evaluation window. Both matrices use the same diverging color scale, with zero at its midpoint. The engine detail uses a separate horizontal scale, common to both ETF sets, to show the small positive and negative differences. All marks are point estimates.}
\end{center}

\clearpage
\subsection{Exposure and trading}

Table~\ref{tab:expansion-trading} connects exposure and trading activity to the direct return deduction for costs. We sum the value of purchases and sales as fractions of portfolio value immediately before each trade. This includes both sides of a rotation and changes in total risky exposure. The cost column is $252\,\operatorname{mean}(d_t)$, where $d_t=r_t^{\mathrm{gross}}-r_t^{\mathrm{net}}$. Both returns use the previous close's portfolio value as their denominator. This measures charges along the baseline path. A rerun with zero charges changes subsequent wealth and trade sizes as well. The timing of holdings, trades, and valuation determines that distinction \citep{boyd2017trading}.

\begin{center}
  \parbox{\textwidth}{\refstepcounter{table}\label{tab:expansion-trading}%
  \small Table~\thetable: Baseline mean risky exposure, annual purchases plus sales as a multiple of portfolio value, and annualized mean cost deduction. For the regression model, each entry averages the same metric across five seeds. All paths use the same charge of 13 basis points on purchases and sales.}
  \smallskip
  \setlength{\tabcolsep}{4.5pt}
  \begin{tabular}{@{}llrrr@{}}
    \toprule
    Fund set & Strategy & \shortstack{Mean risky\\exposure} & \shortstack{Annual purchases\\plus sales} & \shortstack{Annualized\\cost deduction} \\
    \midrule
    \ExpansionTradingRows
    \bottomrule
  \end{tabular}
\end{center}

In both ETF sets, mean annual purchases plus sales relative to portfolio value are more than twice as large for gradient boosting as for cross-sectional momentum. Gradient boosting also incurs larger cost deductions at the same proportional charge. In countries it also has lower mean risky exposure. These differences show why comparing return or Sharpe includes the consequences of exposure, allocation, and trading together.

\end{document}